\documentclass[10pt,twocolumn,letterpaper]{article}

\usepackage{wacv}              

\usepackage{graphicx}
\usepackage{amsmath}
\usepackage{amssymb}
\usepackage{booktabs}
\usepackage{times}
\usepackage{epsfig}
\usepackage{graphicx}
\usepackage{amsmath}
\usepackage{amssymb}
\usepackage{booktabs}
\usepackage{algorithm}
\usepackage{algorithmic}
\usepackage{comment}
\usepackage[dvipsnames]{xcolor}
\usepackage{multirow}
\usepackage{colortbl}
\definecolor{mygray}{gray}{.9}

\newcommand\blfootnote[1]{%
  \begingroup
  \renewcommand\thefootnote{}\footnote{#1}%
  \addtocounter{footnote}{-1}%
  \endgroup
}

\def\XX{{\mathbf{X}}}
\def\ZZ{{\mathbf{Z}}}
\def\yy{{\mathbf{y}}}
\def\softmax{{\mathbf{\hat y}}}
\def\sD{{\mathcal{D}}}
\def\real{{\mathbb{R}}}

\usepackage{pifont}
\newcommand{\xmark}{\ding{55}}
\newcommand{\cmark}{\ding{51}}
\usepackage{xcolor}

\newcommand{\pascal}{\mbox{PASCAL VOC2012}}

\usepackage[includeheadfoot,margin=2cm]{geometry}
\usepackage[font=small,labelfont=bf,tableposition=top]{caption}
\usepackage{blindtext}
\usepackage{mmstyles}

\usepackage[pagebackref,breaklinks,colorlinks]{hyperref}

\usepackage[capitalize]{cleveref}
\crefname{section}{Sec.}{Secs.}
\Crefname{section}{Section}{Sections}
\Crefname{table}{Table}{Tables}
\crefname{table}{Tab.}{Tabs.}

\def\wacvPaperID{142} 
\def\confName{WACV}
\def\confYear{2024}

\begin{document}

\title{Prompting classes: Exploring the Power of Prompt Class Learning \\ in Weakly Supervised Semantic Segmentation}

\author{Balamurali Murugesan*\\
ETS Montreal\\
\and
Rukhshanda Hussain*\\
Jadavpur University \\ 
\and
Rajarshi Bhattacharya* \\
Jadavpur University \\
\and
\and
Ismail Ben Ayed \\
ETS Montreal \\
\and
Jose Dolz \\
ETS Montreal \\
}

\twocolumn[{%
\renewcommand\twocolumn[1][]{#1}%
\maketitle
\vspace{-10mm}
\begin{center}
    \centering
    \captionsetup{type=figure}
    \includegraphics[width=0.9\textwidth]{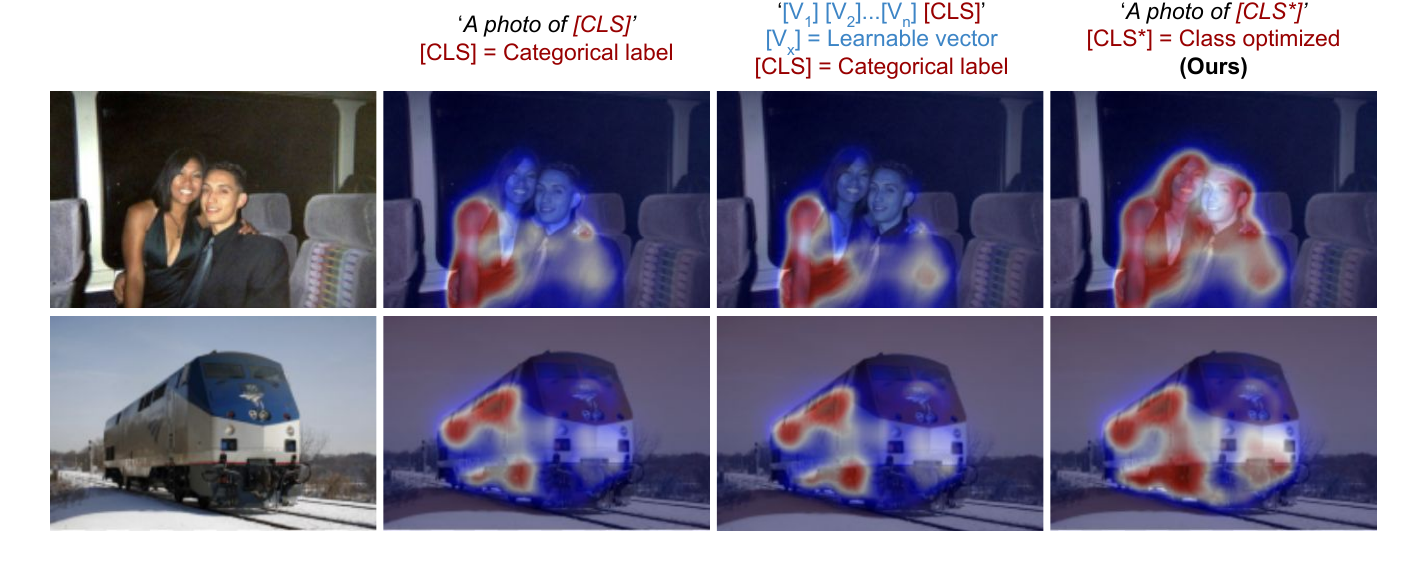}
    \captionof{figure}{\textbf{Impact of the input text prompt on the generation of class activation maps (CAMs).} Employing the ground truth categorical label as [CLS] token (\textit{second column}) does not necessarily result in the best initial CAMs. Furthermore, even though complex techniques to optimize the [CTX] tokens, such as CoOp \cite{zhou2022conditional} (\textit{third column}) may improve the CAMs, we have observed that simply modifying the ground truth class in the [CLS] token by a higher correlated synonym leads to improvements in the identified class-related regions (\textit{fourth column}). }  
    \label{fig:main}
\end{center}%
}]

\thispagestyle{empty}

\begin{abstract}

Recently, CLIP-based approaches have exhibited remarkable performance on generalization and few-shot learning tasks, fueled by the power of contrastive language-vision pre-training. In particular, prompt tuning has emerged as an effective strategy to adapt the pre-trained language-vision models to downstream tasks by employing task-related textual tokens. Motivated by this progress, in this work we question whether other fundamental problems, such as weakly supervised semantic segmentation (WSSS), can benefit from prompt tuning. Our findings reveal two interesting observations that shed light on the impact of prompt tuning on WSSS. First, modifying only the class token of the text prompt results in a greater impact on the Class Activation Map (CAM), compared to arguably more complex strategies that optimize the context. 
And second, the class token associated with the image ground truth does not necessarily correspond to the category
that yields the best CAM. 
Motivated by these observations, we introduce a novel approach based on a \textbf{P}r\textbf{O}mpt c\textbf{L}ass l\textbf{E}arning (\textbf{POLE}) strategy. Through extensive experiments we demonstrate that our simple, yet efficient approach achieves SOTA performance in a well-known WSSS benchmark. These results highlight not only the benefits of language-vision models in WSSS but also the potential of prompt learning for this problem. \blfootnote{$^*$ Authors contributed equally to this article. R. Hussain and R. Bhattacharya did this work as part of a research internship at ETS Montreal.}The code is available at \href{https://github.com/Ruxie189/WSS_POLE}{code link}

\end{abstract}

\section{Introduction}
Image semantic segmentation is a fundamental problem in computer vision, as it serves as a precursor of many tasks, such as medical image analysis or autonomous driving. Fueled by the advances in deep learning, semantic segmentation has experienced a tremendous progress. Nevertheless, obtaining precise pixel-wise annotations is a labor-intensive and time-consuming task. 

To alleviate the annotation burden, weakly supervised semantic segmentation (WSSS) has emerged as an appealing alternative, 
where labels typically come in the form of image tags \cite{fan2020learning,kolesnikov2016seed,hou2018self,lee2019ficklenet}, bounding boxes \cite{song2019box,khoreva2017simple}, scribbles \cite{lin2016scribblesup,tang2018regularized} 
or global constraints \cite{pathak2015constrained,kervadec2019constrained}, among others. In particular, image-level WSSS has received significant attention, as it offers a cost-effective alternative to pixel-level annotations (e.g., 20 seconds reported in \cite{bearman2016s}). Under this setting, WSSS commonly leverages class activation maps (CAMs)  \cite{zhou2016learning} obtained from image classification networks to localize objects. Specifically, these maps are later used as pixel-wise pseudo-labels to train a segmentation model, mimicking full supervision. However, CAMs tend to highlight discriminative regions, while ignoring other useful cues, which results in suboptimal pseudo-labels that do not cover the whole extent of the target objects. 
Narrowing down the existing gap between classification and segmentation tasks is therefore crucial for the progress of WSSS models. To solve this issue, existing approaches intend to complete generated CAMs by forcing the network to focus on more non-discriminative regions, which can be achieved by region mining strategies \cite{kweon2021unlocking,hou2018self}, 
or integrating iterative processes \cite{ahn2018learning}. Despite employing complex CAM refinement strategies, sometimes involving multiple training steps, existing approaches still exhibit suboptimal performance in terms of both completeness of the target objects and segmentation accuracy.

This motivates the exploration of complementary learning strategies that 
can further improve the segmentation performance of these models. 
Vision-language pre-training (VLP) models, such as the recently introduced Contrastive Language-Image Pre-training (CLIP) \cite{radford2021learning} strategy, have the potential to bring WSSS approaches to the next level, as it can associate much wider visual concepts in an image with their corresponding text labels in an open-world scenario. This contrasts with standard WSSS settings, where the fixed set of predetermined object categories limits the quality of generated CAMs due to unnecessary background activations from class-related background pixels. For example, CLIMS \cite{xie2022clims} exposed these issues showing that background pixels related to the class \textit{`railroad'} contributed to the prediction of the CAM associated to the category \textit{`train'}, leading to over-segmented CAMs.

With the rise of these powerful vision-language pre-training models, recent evidence has highlighted the importance of their text input,
 typically referred to as \textit{prompt}, 
 in adapting these models to downstream tasks and datasets. For instance, Zhou \textit{et al.} \cite{zhou2022learning} empirically demonstrated that the use of \textit{'a} [CLS]' or \textit{'a photo of a} [CLS]' as a prompt 
led to substantial differences in the classification performance of the model. Following these findings, recent literature has focused on tuning the context of these prompts, 
typically as continuous learnable vectors \cite{zhou2022learning,zhou2022conditional,ju2022prompting}. Despite its potential importance, the impact of modifying the [CLS] token has been largely overlooked in the context of prompt learning. Additionally, while prompt learning has shown promising results in fine-tuning and classification tasks such as zero-shot image recognition, its effectiveness on other visual recognition problems is not well-understood.



Based on these observations, we explore in this work how vision-language pre-training can be further leveraged to improve the performance of WSSS models. In particular, we want to address the following questions: \textcircled{1} \textit{Is prompt learning useful in weakly supervised segmentation?}, \textcircled{2} \textit{Which parts of the prompt have a greater impact on the generated CAMs?} \textcircled{3} \textit{Can we devise a simple yet effective 
alternative to improve the segmentation performance under the weakly-supervised learning paradigm?}



\textbf{Our contributions} can be summarized as follows:

\begin{itemize}
    \item We provide empirical evidence that modifying the input prompt in VLP models has a direct impact on the generated CAMs in a weakly supervised segmentation scenario, which in turn affects the performance of the segmentation network.

    \item More interestingly, our findings reveal that replacing the [CLS] token in the input prompt has a greater impact on the performance than modifying the prompt context, which contrasts with recent observations in classification problems (See Fig. \ref{fig:main}). Furthermore, the [CLS] token associated with the actual image ground truth does not necessarily correspond to the category that yields the best CAM, and the performance varies considerably across closely related categories. These insights shed light on the importance of careful prompt design in optimizing the performance of segmentation models trained under the weakly-supervised paradigm.

    
    \item Based on these observations, we propose a simple yet efficient strategy to leverage language driven models in the challenging task of WSSS. The resulting model, based on a \textbf{P}r\textbf{O}mpt c\textbf{L}ass l\textbf{E}arning (\textbf{POLE}) approach, learns the category name that produces the highest correlation between the image and a corresponding text prompt, and uses it to further leverage the segmentation performance. 
    
    \item Following the literature, we conduct extensive experiments on PASCAL VOC 2012 to well demonstrate the superiority of our method over other state-of-the-art methods for WSSS.
    
    
\end{itemize}

\section{Related Work}


\noindent \textbf{Weakly supervised semantic segmentation.} Due to its low annotation cost, WSSS based on image-level labels has gained increasing popularity. These methods rely on class activation maps (CAMs) to identify target object regions by discovering informative pixels for the classification task. As discovered regions are typically highly discriminative and fail to cover the whole context of the target objects, recent literature focuses on generating high-quality CAMs by refining initial estimations from simple models. A common strategy is to mine or erase regions at either image \cite{wei2017object,zhang2021complementary,kweon2021unlocking} or features level \cite{hou2018self,lee2019ficklenet}, and can be seen as a way of preventing a classifier from focusing exclusively in highly discriminative areas. Other works have instead exploited sub-categories dependencies \cite{chang2020weakly}, cross-image semantics \cite{fan2020cian,sun2020mining}, attention mechanisms \cite{wu2021embedded}, equivariant constraints \cite{wang2020self,patel2022weakly} and pairwise semantic affinities \cite{ahn2018learning,wang2020weakly}. 
Furthermore, additional supervision, such as saliency maps can be also integrated to provide additional hints about the location of the target object \cite{lee2021railroad,jiang2022l2g}. More recently, visual transformers (ViT) \cite{dosovitskiy2020image} have been also leveraged to improve original CAMs \cite{xu2022multi,li2022transcam}, demonstrating superior performance than their CNNs counterparts. 






\noindent \textbf{Contrastive Language-Image Pre-training (CLIP) based semantic segmentation.} 
Very recently, large-scale VLP models, such as CLIP \cite{radford2021learning}, have demonstrated to improve significantly the performance of vision models on classic recognition tasks, such as zero-shot object detection \cite{gu2021zeroshot}, few-shot learning \cite{hu2022pushing} and zero-shot semantic segmentation \cite{li2021language,zhou2023zegclip}. Closely related to our work, CLIMS \cite{xie2022clims} integrates CLIP in the context of weakly-supervised segmentation, which enhances the initial CAMs by highlighting more comprehensive object regions, while suppressing closely-related background areas. Inspired by the improvement observed in the robustness and generability of visual recognition models driven by language assistance, our work delves deeper into understudied factors, particularly in the weakly-supervised scenario. We stress that our work is different from \cite{xie2022clims}. In particular, CLIMS \cite{xie2022clims} proposes to leverage standard CLIP in WSSS, whereas we further explore the effect of the given prompt on this task. As we will show in our empirical validation, properly designing the input prompt results in significant improvements over the standard text prompts. More surprisingly, \textit{using the class ground truth as categorical name in the input text prompt does not necessarily yields the best segmentation results.} 



\noindent \textbf{Prompt learning} in visual recognition problems is a rapidly growing research direction, whose popularity stems from the promising results observed in NLP tasks \cite{lester2021power,li2021prefix,liu2023pre,raffel2020exploring}. For example, recent works in prompt learning \cite{du2022learning,rao2022denseclip,zhou2022conditional,zhou2022learning,wang2022learning,nayak2022learning} have achieved promising results on several vision-language tasks, notably in classification. In addition to tackle a different task, i.e., classification \textit{vs.} weakly supervised segmentation, the main differences with our work is that these approaches mostly study prompt learning from a context perspective. In particular, existing literature considers the class token [CLS] as a fixed word embedding, while optimizing the context \cite{du2022learning,rao2022denseclip,zhou2022conditional,zhou2022learning,wang2022learning} or attributes \cite{nayak2022learning}. In most cases, the context tokens are represented by learnable continuous vectors, which yield to text embeddings lacking semantic knowledge \cite{zhou2022learning}. In contrast, our approach performs a selection on a finite set of potential synonyms, which facilitates both the search and the interpretation of the selected token. 



\begin{figure*}[t]
    \centering
    \includegraphics[width=\linewidth]{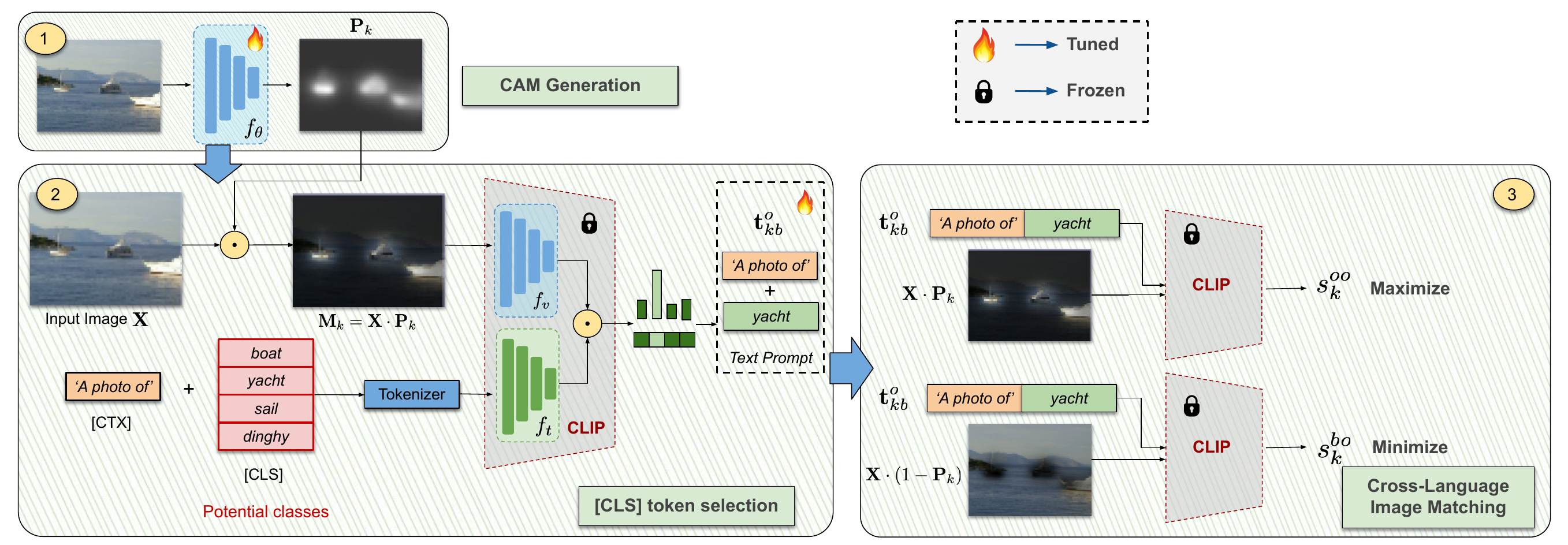}
    \caption{\textbf{Proposed Weakly Supervised Segmentation approach.} \textbf{1)} Class activation maps are generated for an input image $\mathbf{X}$. \textbf{2)} 
    CLIP pre-trained visual and text encoders ($f_{\theta}$ and $f_{\theta}$) are leveraged to find the category name [CLS] presenting the highest correlation with the image $\mathbf{M}_k$, the result of multiplying the input image $\mathbf{X}$ and its corresponding CAM $\mathbf{P}_k$. \textbf{3)} With the [CLS] token selected, we generate the input text prompt $\textbf{t}^o_{kb}$ to the Cross-Language Image Matching (CLIMS) learning framework.}
    \label{fig:motivation_clims}
\end{figure*}

\section{Methodology}
\label{sec:method}


\subsection{Problem setting.} Let us denote $\sD=\{(\mX_i,\yy_i)\}^N_{i=1}$ as a weakly labeled dataset, where $\mX_i \in \real^{\Omega_i}$ is an input image, $\Omega_i$ denotes its spatial dimensionality, $\yy_i \in \{0,1\}^K$ its associated one-hot encoded image label\footnote{In PascalVOC, multiple classes can be present in the same image, where $\yy$ becomes a multi-class one-hot encoded vector.}, and $K$ indicates the number of categories. 
Thus, the goal of weakly supervised semantic segmentation is to provide pixel-wise predictions from an input image $\XX_i$ given its corresponding image-level label $\yy_i$.


\subsection{Our framework.}
\label{sec:framework}


\noindent \textbf{Preliminaries: Class Activation Maps.} 
We first revisit the generation of class activation maps (CAM) from the image-level labels, a popular strategy in WSSS. Let us first define a feature extractor $f_{\theta}(\cdot)$, which can be represented by a deep neural network parameterized by $\theta$. Thus, for a given image $\XX$, the feature extractor provides a representation $\ZZ \in \real^{C \times H' \times W'}$, where $C$ is the number of channels and $H'$ and $W'$ represent the dimensionality of the feature map. 
To provide CAMs, a global average pooling (GAP) layer, followed by a $1 \times 1$ convolutional layer $\mW \in \mathbb{R}^{C\times K}$ is applied to the learned features $\ZZ$ from an image $\XX$. Then, the resulting logits are mapped into probabilities  $\softmax \in [0,1]^K$ by applying a sigmoid function. 
To train the neural network, we follow the literature \cite{xu2021leveraging,xu2022multi,xie2022clims} and use the multi-label soft-margin loss as the classification function:



\begin{equation}
     \mathcal{L}(\softmax, \yy)= -\frac{1}{K}\sum_{k=1}^K (y^k \log \hat{y}^k + \left(1-y_k\right) \cdot \log \left(1-\hat{y}_k\right)). 
     \end{equation}

Once the backbone network is trained, the initial CAMs can be obtained as follows:
\begin{equation}
	\mP_k(h,w) = \mW_k^\top\mZ(h,w), \label{eq:cam}
\end{equation}
where $\mP_k$ is the activation map for a given category $k$.

\noindent \textbf{Learning objectives.} The framework used for this work shares the same overall structure as the recent CLIMS \cite{xie2022clims}, as it represents the first weakly-supervised segmentation approach integrating CLIP text embeddings. Nevertheless, we stress that in our work we study how we can leverage prompt learning to further improve the performance of weakly supervised segmentation models. 
In particular, the standard GAP layer employed to generate CAMs is replaced by a sigmoid function $\sigma(\cdot)$, resulting in $\mP_k(h,w)= \sigma(\mW_k^\top\mZ(h,w))$. The pretrained CLIP model \cite{radford2021learning} uses a visual and a text encoder which we denote as: $f_v$ and $f_t$, respectively. Instead of passing the raw input image $\XX$ to the CLIP image encoder, it is multiplied by $\mP_k$, with the goal of focusing only on the class highlighted by its corresponding CAM. Moreover, to avoid interferences from related background regions, the input image is also multiplied by $(1-\mP_k)$. Hence, we can create two different visual embeddings: the embedding of the target category ($\vv_k^{io}$), and its background ($\vv_k^{ib}$), which are formally given by:

\begin{equation}
	\vv_k^{io} = f_v(\mX \cdot \mP_k ),\quad \vv_k^{ib} = f_v(\mX \cdot  (1-\mP_k) ).
\end{equation}

Now, for all the potential object classes $k$ and their corresponding text inputs $\vt^o_k$, we obtain their text embeddings, which are referred to as $\vv_k^{to}$:

\begin{equation}
    \vv_k^{to} = f_t(\vt^o_k).
\end{equation}

Note that to generate these embeddings, we just need to provide the different text inputs to the trained CLIP text encoder $f_t(\cdot)$. Following the reasoning behind the training of CLIP,  the foreground image embedding $\vv_k^{io}$ should be highly correlated to the text embedding $\vv_k^{to}$ of that particular class. In contrast,  the background image embedding $\vv_k^{ib}$ should have a much lower correlation with the object classes.  This can be modeled by using the following objective function: 
\begin{equation}
\label{eq:cont}
	\mathcal{L}_{Cont} = -\alpha \sum_{k=1}^{K}  y_k \cdot \log (s^{oo}_k) - \beta \sum_{k=1}^{K} y_k \cdot \log (1 - s^{bo}_k),
\end{equation}

where $s^{oo}_k=\text{sim}(\vv^{io}_k, \vv^{to}_k)$ and $s^{bo}_k=\text{sim}(\vv^{ib}_k, \vv^{to}_k)$ represent the \textit{object-to-object} and \textit{background-to-object} similarities between visual and text embeddings,  computed as a cosine similarity. 
Both terms in Eq.  \ref{eq:cont} act together to ensure that the activation map $\mP_k$ covers the maximum possible region of the target,  while excluding related background.

 \subsection{Category and image-driven prompt generation}

\noindent \textbf{Finding potential category related embeddings.} To generate the object text representation $\vv^{to}_k$ for a class $k$, the standard input prompt given to the text encoder has the following format: a context token [CTX] followed by the class name token [CLS] and ended by a punctuator ('.'). While the literature in prompt learning for large-scale visual language pre-trained models focuses on learning the context [CTX] \cite{zhou2022conditional}, CLIMS \cite{xie2022clims} uses a fixed prompt, where the [CLS] token corresponds to the categorical label of the image. Contrary to these works, we hypothesize that modifying the input text prompts by optimizing only the [CLS] token has a greater impact on the generated CAMs. Indeed, as we will show empirically in the results section, using the ground truth class as a [CLS] token does not necessarily always results in the best segmentation performance.


Let us suppose that we take an input image $\XX$ with its corresponding image class label $\yy$, which indicates the $k$ categories present on the image. 
For each category $k$ in $\yy$, we obtain a set of similar words, in terms of closeness in the semantic space, using chatGPT \cite{chatgpt}. More concretely, we provide the following query as input to chatGPT \textit{"Give me $m$ semantically similar words for [CLS] and also print the cosine similarity scores based on CLIP model"}, where [CLS] is a class name. This returns a list of $m$ words along with their similarity scores for that particular [CLS]. This means that for each class [CLS], we can derive a set of $m$ closest words, denoted as $\mathcal{S}=$[CLS$_1$,CLS$_2$,...,CLS$_m$]. With this set of related categories, we can create a set of $m+1$ potential text prompts $\mathcal{T}=[\vt^o_{k0}, \vt^o_{k1},\vt^o_{k2}, \dots \vt^o_{km}]$, where $\vt^o_{k0}$ is the text prompt containing the categorical ground truth label for class $k$, i.e., {[CLS]} followed by a fixed [CTX] token, and $\vt^o_{k1},...,\vt^o_{km}$ are composed of the fixed [CTX] followed by a variable [CLS] token chosen from set $\mathcal{S}$. 
Now, we can extract an embedding for each of the prompts in $\mathcal{T}$ from the CLIP text encoder, resulting in $\mathcal{V}=[\vv^{to}_{k0}, \vv^{to}_{k1},\vv^{to}_{k2},\dots\vv^o_{km}]$. 

\noindent \textbf{How to select the best [CLS]?} 
Given the input image $\XX$ and its generated class activation map $\mP_k$ for class $k$, we can obtain an image focusing on discriminative regions for that category by simply doing $\mM_k = \mX \cdot \mP_k$
The resulting image, $\mM_k$, is given to the CLIP image encoder to get a compressed representation of $\XX$, i.e., $\vv_k^{io}$. similar to the steps described in Section \ref{sec:framework}. In contrast, we now obtain a correlation between each of the closest words in set $\mathcal{S}$ for the class $k$ and the CAM activated image $\mM_k$. In particular, this correlation is found by computing a similarity score between the visual and text embeddings: $\vv_k^{io}$ and every $\vv^{to}_{kj}$ for $j \in \{0,1,2,3, \dots, m\}$, given as:

\begin{equation}
    \text{sim}(\vv_k^{io}, \vv^{to}_{kj}) = \frac{\vv_k^{io} \cdot \vv^{to}_{kj}}{|\vv_k^{io}| |\vv^{to}_{kj}|},
\end{equation}

which generates a vector containing the similarities between the visual encoding and each of the text encodings $[s_{k0},s_{k1},\dots, s_{km}]$. From this similarity vector, we select the most correlated [CLS] token, which corresponds to the text embedding $\vv^{to}_{kj}$ with the highest similarity with $\vv_k^{io}$,  computed with the $\operatorname{argmax}$ operator.

\subsection{Weakly supervised adaptors}

Following the success of adaptors in pre-trained language-vision models for classification tasks \cite{rao2022denseclip,zhang2022tip,gao2021clip}, we propose to further improve our segmentation network by integrating image and text adaptors. In particular, and similar to \cite{gao2021clip} in classification, we introduce two MLP layers $A_v(\cdot)$ and $A_t(\cdot)$ to transform the embeddings in the image side and text space, respectively, which is formulated as:

\begin{align}
\label{eq:adapter}
    \vv_k^{io*} &= \vr_v \cdot A_{v}(\vv_k^{io}) + (1 - \vr_v) \cdot \vv_k^{io} \\
    \vv_k^{to*} &= \vr_t \cdot A_{t}(\vv_k^{to}) + (1 - \vr_t) \cdot \vv_k^{to},
\end{align}

with $A_v(\vv_k^{io})=ReLU(\vv_k^{io} \mathbf{W}_1^v)\mathbf{W}_2^v$ and $A_t(\vv_k^{to})= ReLU(\vv_k^{to} \mathbf{W}_1^t)\mathbf{W}_2^t$, where $\mathbf{W}$ represents the learnable parameters of the MLP layers. Furthermore, the parameters $\vr_v$ and $\vr_t$ are learnable vectors of the same shape as the original embeddings, and are used to selectively suppress or amplify the refinement of each feature through the MLP layers.  Note that this contrasts with standard adapters, i.e., \cite{rao2022denseclip,zhang2022tip,gao2021clip}, whose balancing weight is a fixed hyperparameter. Our hypothesis is that using a fixed scalar to control the importance of each embedding is suboptimal, as the features refinement process may differ across images as well as depend on the class variation of the dataset. 



\section{Experiments}
\subsection{Experimental Settings}

\noindent \textbf{Dataset and evaluation protocol.}
Following CLIMS \cite{xie2022clims}, as well as other recent WSSS works \cite{lee2021anti}, we conduct experiments on the popular PASCAL VOC 2012 \cite{everingham2009pascal} benchmark. This dataset contains images with 20 foreground classes, which are split into 1,464 for training, 1,449 for validation and 1,456 for testing. The training set is augmented with 10,582 images and their associated image-level annotations from SBD \cite{hariharan2011semantic}. 
To evaluate the performance of the proposed method, we resort to the mean intersection over union (mIoU). Last, while the results reported in the ablation studies are obtained on the training set, the results for the validation and testing sets of PASCAL VOC are obtained from the official evaluation server. 


\noindent \textbf{Implementation Details.} We followed the default settings of CLIMS \cite{xie2022clims} for training. In particular, input images are randomly rescaled and then augmented by random cropping to $512\times 512$, as well as by horizontal flipping. 
We use SGD as the default optimizer, with a cosine annealing policy for scheduling the learning rate, and a batch size of 16 images. 
The model is trained for 10 epochs, with an initial learning rate of 0.00025 and a weight decay of 0.0001. We follow ~\cite{affinitynet} to adopt ResNet-50~\cite{he2016deep} as backbone network for the generation of initial CAMs. All models are implemented in PyTorch and trained on NVIDIA A100 GPU with 40 GB memory. Furthermore, as the initial CAMs coarsely covers the target object, we further perform a refinement step with IRNet~\cite{ahn2019weakly}, to improve their quality before using them as pseudo ground-truth masks, a common practice in WSSS. 

%


\subsection{Results}

\noindent \textbf{How effective is prompt learning for weakly supervised segmentation?} In this section we assess whether modifying the input text prompt leads to performance differences, which corresponds to our first question. 
To do this, we first manually selected two popular prompts: ``\textit{A photo of} [CLS].", and ``\textit{An image of} [CLS].", 
which are employed over all the images of the whole dataset. Note that the original CLIMS \cite{xie2022clims} used the first prompt, and it is therefore considered as the baseline model. Furthermore, inspired by the recent advances on prompt learning, we also evaluate the impact different strategies, which model the different tokens with learnable continuous vectors: CoOp \cite{zhou2022learning} ([CTX] tokens), \textit{target optimization} baseline in \cite{wang2022learning} ([CLS] tokens) and a modified version of DeFo \cite{wang2022learning}. These results, which are reported in Table \ref{table:prompts}, reveal that the text input, i.e., prompt, of the pre-trained vision-language model plays an important role on the segmentation performance.  Indeed, we can observe that depending on the prompt employed, performance differences may diverge up to 3\%, particularly on the final generated CAM (\textit{last column}). 


\begin{table}[h!]
	\centering
	\footnotesize
	\caption{\textbf{Does prompt learning improve the performance of weakly supervised segmentation?} Comparison of the quality of initial CAMs and refined pseudo ground-truth masks obtained by different prompt learning strategies (with R50 as backbone), where either [CTX] or [CLS] tokens are modified. Evaluation is reported on the \textit{train} set of \pascal, and refinement of the pseudo-masks is performed using \textbf{RW} (IRN \cite{ahn2019weakly}). [CTX] and [CLS] are used to indicate which part of the prompt is optimized in each approach. Furthermore, in the approaches optimizing the [CLS] token, 'V' indicates a continuous learnable vector, whereas CLS* represents the class selected among a set of potential classes. }
	\label{tab:cam_rw}
	\resizebox{\linewidth}{!}{
		\begin{tabular}{l c c l  c c}
			\toprule
			Method & [CTX] & [CLS] & Prompt  & CAMs & +\textbf{RW}\\
			\midrule  
Manual \cite{xie2022clims} & \cmark & \xmark & ``\textit{A photo of} [CLS]."  &56.6 & 70.5 \\
Manual &\cmark	& \xmark &``\textit{An image of} [CLS]."  &  56.5 & 71.0 \\

CoOp \cite{zhou2022conditional} & \cmark & \xmark & [V]$_1$[V]$_2\dots$[V]$_N$[CLS]. &  57.6 & 73.1 \\ 
DeFo \cite{wang2022learning} & \cmark & \cmark  &[V]$_1$[V]$_2\dots$[V]$_N$[$V_{CLS}$] &  56.6 & 73.2  \\
Target optimization & \xmark & \cmark &"\textit{A photo of} [$V_{CLS}$]."  & 56.8 & 73.1  \\
Ours  & \xmark & \cmark &"\textit{A photo of} [{CLS}*]."  & \bf 58.7 & \bf 73.6  \\
\bottomrule
\end{tabular}}
\label{table:prompts}

\end{table}

\begin{table}[t!]
	\centering
	\caption{Comparison of the quality of initial CAMs and refined pseudo ground-truth masks using \textbf{RW} (PSA~\cite{affinitynet}) on PASCAL VOC2012. The mIoU values here are reported on the \textit{train} set. Backbone denotes the backbone network to generate CAMs. Best approaches (CAM and refined CAM) highlighted in bold.}  
	\label{tab:cam_rw}
	\vspace{-10pt}
	\resizebox{\linewidth}{!}{
		\begin{tabular}{l c c c}
			\toprule
			Method & Backbone & CAMs & +\textbf{RW}\\
			\midrule  
			PSA$_\text{ CVPR'2018}$~\cite{ahn2018learning}   & WR38 & 48.0 & 61.0 \\
			SC-CAM$_\text{ CVPR'2020}$~\cite{chang2020weakly}    & WR38 & 50.9 & 63.4 \\
			SEAM$_\text{ CVPR'2020}$~\cite{wang2020self}    & WR38& 55.4 & 63.6 \\
			AdvCAM$_\text{ CVPR'2021}$~\cite{lee2021anti}    & R50& 55.6 & 68.0 \\
            MCTformer$_\text{ CVPR'2022}$~\cite{xu2022multi}  & ViT  & \bf 61.7 & 69.1\\
            SIPE$_\text{ CVPR'2022}$~\cite{chen2022self} & R50 & 58.6 & 64.7\\
            RECAM$_\text{ CVPR'2022}$~\cite{chen2022class} & R50 & 54.8  & 70.5 \\
            AdvCAM+W-OoD$_\text{ CVPR'2022}$~\cite{lee2022weakly} & R50 & 59.1 & 72.1\\
            AFA$_\text{ CVPR'2022}$~\cite{ru2022learning} & MiT & & 68.7 \\
			CLIMS$_\text{ CVPR'2022}$~\cite{xie2022clims} & R50 & 56.6  & 70.5 \\
			VWL$_\text{ IJCV'2022}$~\cite{ru2022weakly} & R101 & 57.3  & 71.4 \\
			 AEFT$_\text{ ECCV'2022}$~\cite{yoon2022adversarial} & WR38 & 56.0  & 71.0 \\
			 ViT-PCM$_\text{ ECCV'2022}$~\cite{rossetti2022max} &  ViT-B/16 & -- & 71.4 \\
			 ESOL$_\text{ NeurIPS'2022}$~\cite{li2022expansion} & R50 & 53.6  & 68.7 \\
			\rowcolor{mygray}
			\textbf{POLE (Ours)}  &  R50 & 59.0 & \textbf{74.2}  \\

			\bottomrule
	\end{tabular}}
\end{table}

\noindent \textbf{Context \textit{vs.} category in prompt learning.} Once we have observed empirically that modifying input prompts results in performance differences, one question that naturally arises is which component of the prompt must be changed. Table \ref{table:prompts} shows that replacing a standard [CTX] token (i.e., \textit{'A photo of'}) by a similar sequence (i.e., \textit{'An image of'}) brings 0.5\% difference. Nevertheless, if the [CTX] token is optimized for the whole dataset, e.g., CoOp \cite{zhou2022conditional}, these differences are further increased, with similar results if [CLS] token is optimized as a continuous vector (e.g., DeFo \cite{wang2022learning} and \textit{Target Optimization}). Last, we can observe that only optimizing the [CLS] prompt, based on a set of pre-defined closely-related categories, actually provides the best performance across all the methods. These findings align with recent observations in Natural Language Inference \cite{logan2022cutting}, which suggest that hand-crafted prompts conveying meaningful instructions outperform automatically optimized prompts.

\noindent \textbf{Comparison to state-of-the-art.} Previous experiments empirically demonstrated that modifying the input prompt can significantly improve the performance of weakly supervised segmentation models. These observations motivated the proposed approach, which we benchmark against state-of-the-art models to show its superiority. Note that in what follows, the proposed approach, POLE, is composed of the [CLS] token selection process and the adaptor with learnable weights (eq. \ref{eq:adapter}). First, Table \ref{tab:cam_rw} reports the results of state-of-the-art methods in the generation of pseudo-masks.  We can observe that,  even compared to very recent models, the proposed approach brings substantial improvements, ranging from 2 to 6\%. We interpret that the performance gain observed comes from the highest correlation between the selected category name and the content of the CAM-activated region in the image., which may capture larger discriminant areas of semantic objects. 
More interestingly, the proposed approach also outperforms recent methods that use additional information, for example in the form of extra saliency maps, which are typically trained on a supervised foreground-background detection dataset. Thus, based on these results we can argue that our approach represents an effective alternative to generate initial CAMs.



\begin{table}[h!]
	\centering
	\caption{Evaluation results on PASCAL VOC2012 \textit{val} and \textit{test} sets. The best results are in \textbf{bold}. Sup. denotes the weak supervision type. $\mathcal{F}$ denotes full supervision. $\mathcal{S}$ denotes saliency map supervision. $\mathcal{I}$ denotes image-level supervision. Seg. denotes the segmentation network. Bac. denotes the backbone network for CAMs generation. V1: DeepLabV1. V2: DeepLabV2. V16: VGG-16~\cite{simonyan2014very}. R50: ResNet-50~\cite{he2016deep}. WR38: WideResNet38~\cite{wu2019wider}. Segmentation network pretrained with ImageNet otherwise using MS COCO dataset ($^\ddag$). Approaches based on visual transformers (\textit{last section}) and convolutional neural networks (\textit{before-last section}) are separated, and best method in each is highlighted in \textbf{bold}. 
	}
	\vspace{-3pt}
	\resizebox{\linewidth}{!}{
		\begin{tabular}{c l l c c c}
			\toprule
			Sup. &Method & Seg. & Bac. & \textit{val} & \textit{test}\\
			\midrule
			\multirow{2}{*}{$\mathcal{F}$}&DeepLabV2$_\text{ TPAMI'18}$~\cite{deeplabv2}  &  -  & - & 77.6 &  79.7 \\
			&WideResNet38$_\text{ PR'19}$~\cite{resnet38}  &  -  & - &  80.8 &  82.5 \\
			\midrule
			\multirow{6}{*}{$\mathcal{I+S}$}&NSROM$_\text{ CVPR'21}$~\cite{nsrom} & V2$^\ddag$-R101 & V16 & 68.3 & 68.5 \\
			&DRS$_\text{ AAAI'21}$~\cite{drs} & V2$^\ddag$ & V16 & 70.4 & 70.7 \\
			&EPS$_\text{ CVPR'21}$~\cite{lee2021railroad} & V2$^\ddag$-R101 & WR38 & 70.9 & 70.8 \\
			&EDAM$_\text{ CVPR'21}$~\cite{wu2021embedded}  & V2$^\ddag$-R101 & WR38 & 70.9 & 70.6\\
			&AuxSegNet$_\text{ ICCV'21}$~\cite{AuxSeg}  & WR38 & - & 69.0 & 68.6 \\	
                &SANCE$_\text{ CVPR'22}$~\cite{AuxSeg}  & V2-R101 & R50 & 72.0 & 72.9 \\	
			\midrule
			
			\multirow{16}{*}{$\mathcal{I}$} &SEAM$_\text{ CVPR'20}$~\cite{wang2020self}  & V3-R38  & WR38 & 64.5 &  65.7 \\
			&BES$_\text{ ECCV'20}$~\cite{bes}  & V2-R101  & R50 & 65.7 &  66.6 \\
			&SC-CAM$_\text{ CVPR'20}$~\cite{chang2020weakly}  & V2-R101 & WR38 & 66.1 &  65.9 \\
			&A$^2$GNN$_\text{ TPAMI'21}$~\cite{a2gnn}  & V2-R101  & WR38 &  66.8 &  67.4 \\
			&VWE$_\text{ IJCAI'21}$~\cite{ru2021learning}  & V2-R101  & R50 &  67.2 &  67.3 \\
                &AdvCAM$_\text{CVPR'21}$~\cite{lee2021anti}  & V2-R101  & R50 & 68.1 &  68.0 \\
			   & VWL$_\text{IJCV'22}$~\cite{ru2022weakly} & V2$^\ddag$-R101 & R101 & 70.6  & 70.7 \\
      		& SIPE$_\text{ CVPR'22}$~\cite{chen2022self}  &  V2-R38 & R50 & 68.2 & 69.5 \\
			& CLIMS$_\text{ CVPR'22}$~\cite{xie2022clims}& V2$^\ddag$-R101  & R50 & 70.4 &  70.0 \\
            & AdvCAM+W-OoD$_\text{CVPR'22}$~\cite{lee2022weakly} & V2-R101 & R50 & 69.8 & 69.9\\
             & SIPE$_\text{CVPR'22}$~\cite{chen2022self} & V2-R101 & R50& 68.8 & 69.7\\
            & RECAM$_\text{CVPR'22}$~\cite{chen2022class} & V2-R101 & R50 & 68.5 & 68.4\\
            & AMN$_\text{CVPR'22}$~\cite{lee2022threshold} & V2$^\ddag$R101 & R50 & 70.7 & 70.6 \\
           & Spatial-BCE$_\text{ECCV'22}$~\cite{wu2022adaptive} & V2-R101 & R38 & 68.5 & 69.7 \\
           &ESOL$_\text{NeurIPS'22}$~\cite{li2022expansion} & V2-R101  & R50 & 69.9  & 69.3 \\ 
            \rowcolor{mygray} & \textbf{POLE (ours)} & V2$^\ddag$-R101 & R50 & \bf 71.5 & \bf 71.4 \\
		  \midrule
           \multirow{3}{*}{$\mathcal{I}$} & AFA$_\text{CVPR'22}$~\cite{ru2022learning}  & MiT-B1 & & 66.0 & 66.3 \\
            & MCTformer$_\text{CVPR'22}$~\cite{xu2022multi}  & V1-R38 & DeiT-S & \bf 71.9 & \bf 71.6 \\
            & ViT-PCM$_\text{ECCV'22}$~\cite{rossetti2022max} & V2-R101  & ViT-B/16 & 70.3 & 70.9 \\
            
            \bottomrule
	\end{tabular}}
\label{tab:segmentation_voc}
\vspace{-5pt}
\end{table}

While the quality of the initial CAMs was evaluated in the previous section, we now assess their impact on the semantic segmentation task.
In particular, Table \ref{tab:segmentation_voc} reports the segmentation performance of the proposed approach compare to state-of-the-art methods in the validation and testing sets of \pascal \, dataset. Compared to approaches that resort to the same supervision, our method provides very satisfactory results, ranking second if all the approaches are considered. Nevertheless, it has been found recently that vision transformers provide much better quality CAMs than conventional convolutional neural networks. Thus, if we just consider the approaches that leverage CNNs for the CAM generation, our approach achieves the best performance, with improvement gains ranging from 0.7\% to 3\% compared to very recent methods (e.g., RECAM, SIPE or CLIMS \cite{xie2022clims}). Furthermore, even benchmarking POLE against recent approaches that use additional supervision, e.g., saliency maps, it yields very competitive performance.

\begin{figure}[h!]
    \centering
    \includegraphics[width=1.0\linewidth]{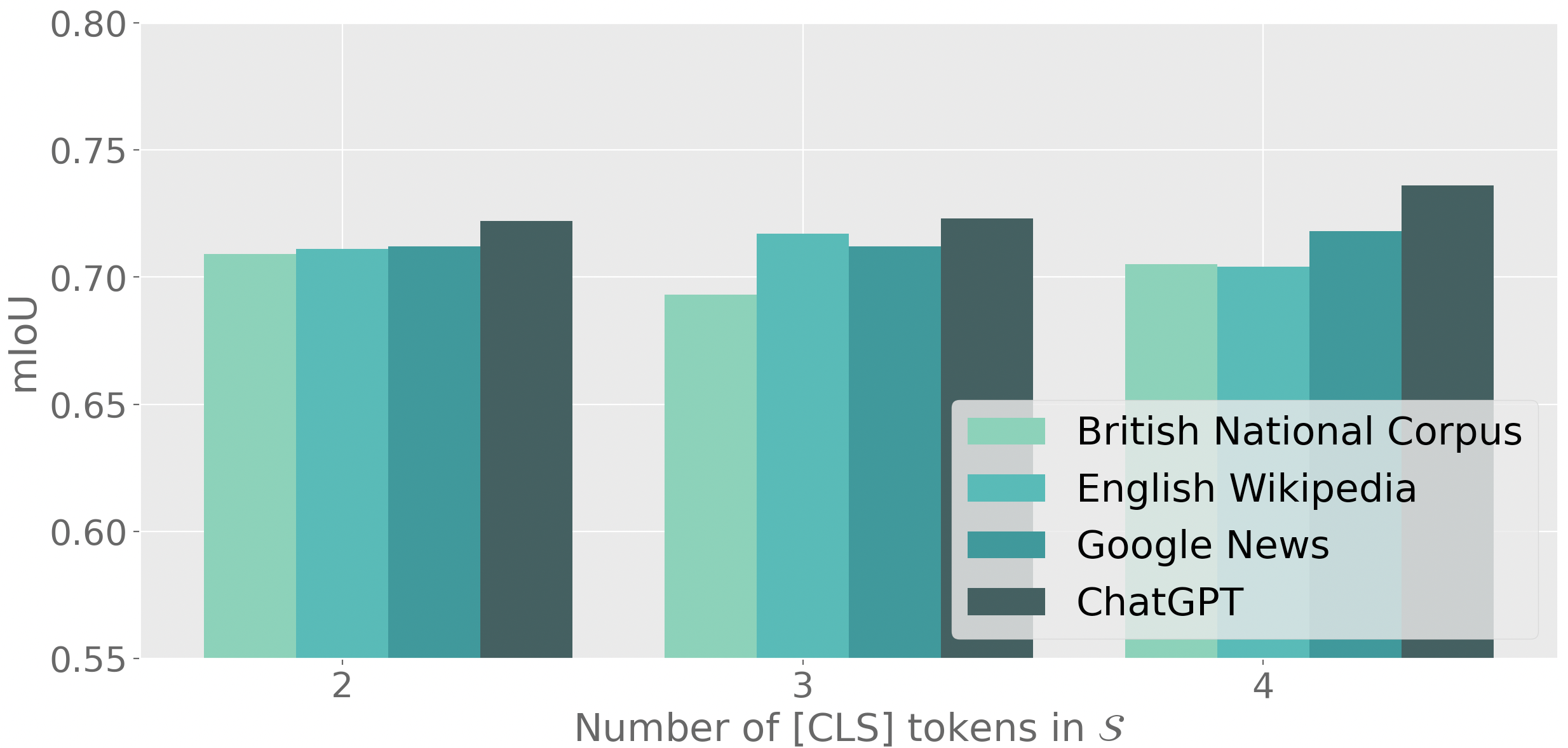}
    \caption{\textbf{Impact of the Corpus choice and number of synonyms selected.} ChatGPT offers the richer variety of synonyms, yielding the best results across other corpus. Furthermore, increasing the number of synonyms (from 
    $2$ up to $4$) further improves the results. Note that the number of synonyms includes the categorical name from the ground truth and the requested close synonyms.}
    \label{fig:ablation}
    \vspace{-2mm}
\end{figure}

\begin{figure}[h!]
    \centering
    \includegraphics[width=0.8\linewidth]{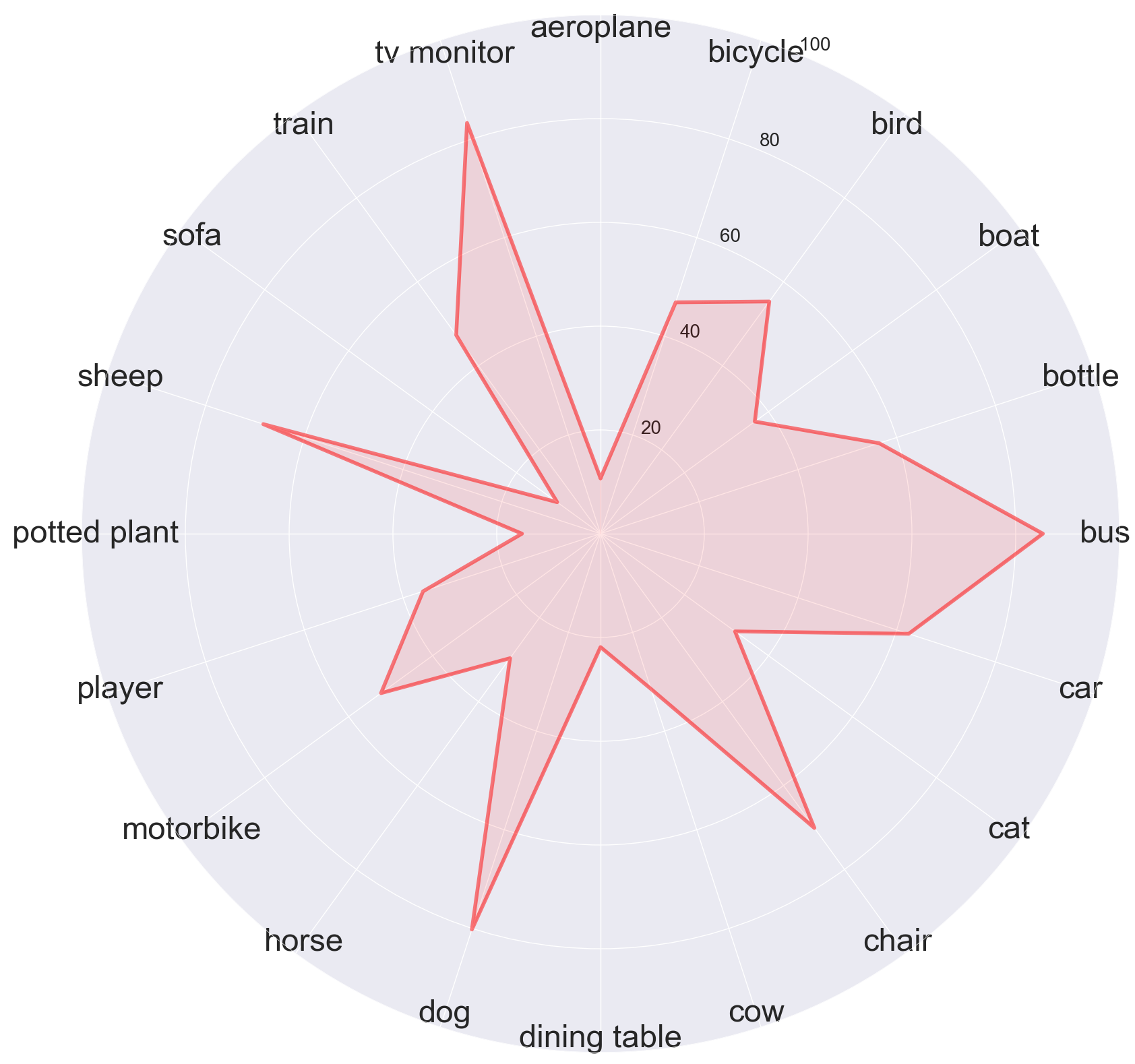}
    \caption{\textbf{What does CLIP think about the best [CLS]? Is the ground truth category chosen everytime? How likely is it that CLIP will select something different?} The plot summarises the percentage of cases where the ground truth category was chosen for an instance of that class. Thus, an inward point on the radial plot indicates that the number of instances where the ground truth category was chosen as the best [CLS] token is considerably low.}
    \label{fig:ablation_pac}
\end{figure}

\noindent \textbf{How many synonyms are sufficient? And from which Corpus?} Previous results have demonstrated empirically that the proposed approach brings substantial improvements by just replacing the categorical name on the ground truth by a closely related synonym. Thus, we now want to evaluate the impact of the corpus selected, as well as the amount of synonyms needed to achieve satisfactory results. In particular, we evaluate the performance of the proposed approach (Fig \ref{fig:ablation}), without adapters, i.e., `A photo of [CLS*].', when the optimal [CLS] token is selected from a set of potential synonyms extracted from different corpus: British National Corpus \cite{bnc2007british}, Google News \cite{kutuzov2017word} and English Wikipedia. The first observation is that, while the use of different corpus increase the performance over the baseline (\textit{`A photo of [CLS].'} in Table \ref{table:prompts}), synonyms from ChatGPT yield the best performance, regardless on the number of names requested. This may be explained by the larger and richer body of text, from a variety of sources, used to train ChatGPT. Next, we can observe that the quality of the generated CAMs typically augments with the number of synonyms (e.g., with ChatGPT we obtain a mIoU of 72.2 \textit{vs} 73.6, from 2 and 4 synonyms, respectively). These results showcase how the most semantically related words, from a natural language standpoint, do not always yield the best performance. Indeed, as the performance increases with the number of synonyms included in our method, one can easily deduce that the synonyms newly added (less correlated than the first ones) may provide better supervisory signals for certain images. Additionally, we investigate the frequency that the actual ground truth category is selected as the [CLS] token, whose results are depicted in Fig. \ref{fig:ablation_pac}. The findings from this radar plots reveal that, surprisingly, the ground truth associated with most instances does not correspond to the most correlated category, which may explain the performance gains observed in our approach. Further exploration on the choice of the synonyms across corpus can be found in Supplemental Material.  

\begin{figure*}[t!]
    \centering
    \includegraphics[width=0.95\linewidth]{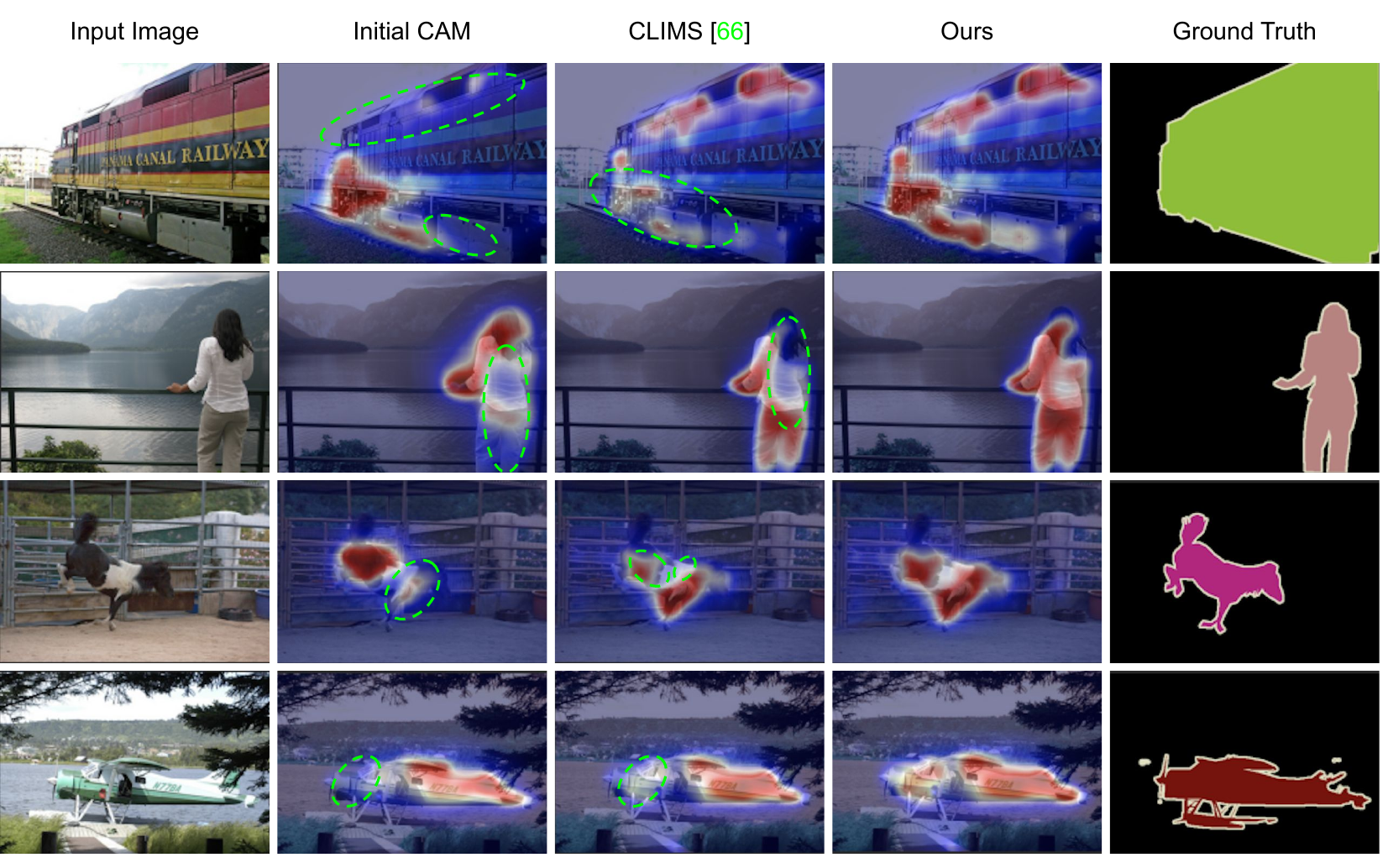}
    \caption{\textbf{Qualitative results of the initial class activation maps.} Green dotted lines ellipses are used to indicate missed regions by previous approaches (original CAMs and CLIMS \cite{xie2022clims}) compared to the proposed method. No refinement on the obtained CAMs is done (e.g., \textbf{RW}) to better illustrate the impact of our approach.}
    \label{fig:main_results}
\end{figure*}

\noindent \textbf{On the impact of the different components.} We observed in Table \ref{table:prompts} that the proposed yet simple strategy to optimize the category name achieves better performance than arguably more complex techniques that attempt to optimize the whole text prompt. In this section, we empirically motivate the use of the proposed adapters, as well as the choice of adding learnable parameters to control the importance of each term in Eq. \ref{eq:adapter}, unlike the fixed hyperparameter used in the existing literature. In particular, table \ref{table:ablationMain} reports these results, where the first observation is that adding the proposed adaptors results in slight improvement compared to the model without them. In contrast, replacing the fixed vectors by learnable ones, the performance is further improved by 0.4\%. Thus, the negligible increase in model complexity due to the adapters, and the performance gain observed, support the choices behind our approach.

\begin{table}[h!]
\vspace{-2mm}
	\centering
	\footnotesize
	\caption{\textbf{Ablation on the main components}. Empirical results that validate the different components of the proposed methodology. A) Image label as [CLS], i.e., CLIMS \cite{xie2022clims}; B) Optimal [CLS] selected; C) Fixed adaptor; D) Learnable adaptor. Results are performed on the \textit{train} set of \pascal.}
    
    
    
	\begin{tabular}{ccccc}
	\toprule
	A & B & C & D & mIoU (\%) \\
	\midrule
    \cmark & & & & 70.5 \\
    \midrule
      & \cmark &  &  & 73.6$_{(+3.1)}$ \\
    
    &  \cmark & \cmark & & 73.8$_{(+3.3)}$ \\
    
   \rowcolor{mygray} &  \cmark &  & \cmark & 74.2$_{(+3.7)}$ \\
			\bottomrule
	\end{tabular}
	
	\label{table:ablationMain}
\end{table}

\noindent \textbf{Qualitative results} of the obtained CAMs, compared to standard CAM generation and the related CLIMS \cite{xie2022clims} are depicted in Figure \ref{fig:main_results}. We can observe that despite CLIMS somehow alleviates the under-segmentation problem in conventional CAMs, it still fails to cover larger target regions (see for example first and second columns). In contrast, POLE typically identifies better larger semantic regions related to the target class, resulting in more complete CAMs compared to related approaches.

\section{Conclusions}
In this work, we have investigated the potential of prompt tuning, an emerging strategy to adapt large pre-trained language-vision models, in the challenging task of weakly supervised semantic segmentation. Our empirical observations have demonstrated that simply replacing the text-token associated with the category name yields better segmentation performance than more complex prompt learning strategies focusing on optimizing the context, which dominate the literature in adapting models. More interestingly, we have observed that employing the corresponding image-level ground truth does not always lead to the best segmentation performance, and closely-related synonyms can indeed result in further performance gains. In light of these findings, we have introduced a simple yet efficient approach, POLE, that selects the most correlated class for a given image in order to generate a better text prompt. Comprehensive experiments have shown that the proposed approach can generate high quality pseudo-labels for WSSS, and achieve state-of-the-art performance in a popular WSSS benchmark.

{\small
\bibliographystyle{ieee_fullname}
\bibliography{main}
}

\clearpage
\setcounter{table}{0}
\setcounter{figure}{0}
\section*{Supplemental Material}
\addcontentsline{toc}{section}{Appendices}
\renewcommand{\thesubsection}{\Alph{subsection}}
\renewcommand{\thetable}{S\arabic{table}}
\renewcommand{\thefigure}{S\arabic{figure}}

\subsection{Extended implementation details}
\label{sup:deeplab}

Here we provide additional implementation details to reproduce the results reported in this work. In particular, batch size was set to 5 and learning rate was fixed at 2.5e-4. Stochastic Gradient Descent was used to optimize the model with a momentum parameter 0.9, and a weight decay of 5e-4. We use a multi-scale testing scheme with the different scales set as 0.5, 0.75, 1.0, 1.25 and 1.5, and the outputs are aggregated using max-pool operations, following recent works \cite{xu2022multi} \cite{xie2022clims}. The choice of the default CRF hyperparameters chosen were guided by \cite{deeplabv2}.

\subsection{Additional details on the prompt learning experiments in \autoref{table:prompts}}
\label{sup:prompt_learning}

\begin{figure*}[!h]
    \centering
    \includegraphics[width=0.95\textwidth]{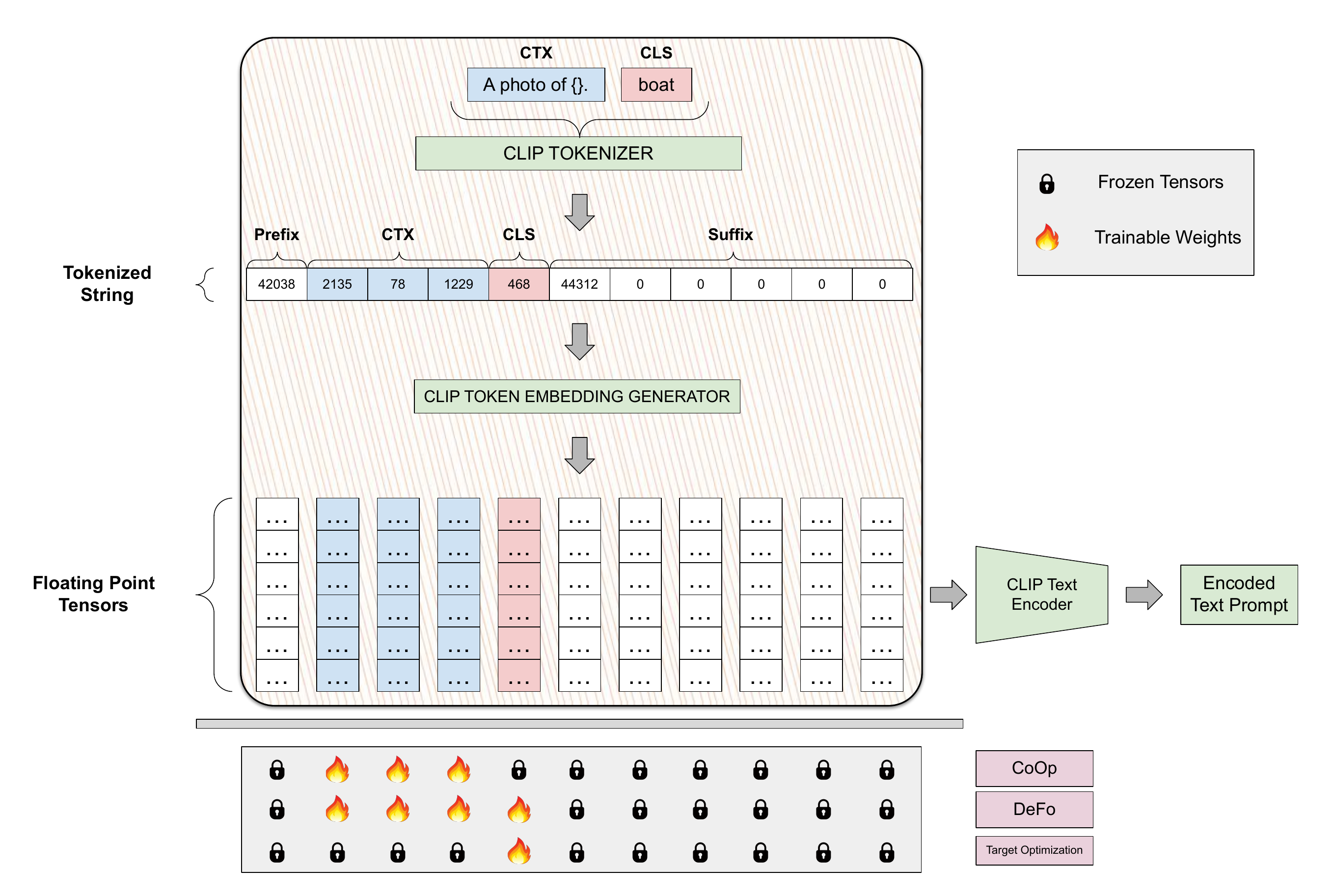}
    \caption{Schematic to show how the text prompts are processed and how each prompt learning technique modifies the process.}
    \label{fig:prompt_learning}
\end{figure*}

\autoref{table:prompts} in the main manuscript showcases the impact of various prompt learning techniques, when implemented for the fine-tuning of our models. In this section we summarize how these approaches were used in our work:

\noindent \textbf{CoOp} \cite{zhou2022learning} uses a trainable context vector of a fixed length ([V]$_1$[V]$_2\dots$[V]$_N$[CLS].), which we set as 3 keeping in mind the length of the default context string: "A photo of". The vector was initialized by tokenizing the same string and encoding it using the pre-trained CLIP \cite{radford2021learning} encoder. The new trainable weights are trained using the same training objectives as in \autoref{eq:cont}. This technique essentially makes the context of the prompt learnable which can be analysed as a possible modification on \cite{xie2022clims}. 

\noindent \textbf{DeFo} \cite{wang2022learning} uses trainable context vectors as well as a trainable class vector ([V]$_1$[V]$_2\dots$[V]$_N$[$V_{CLS}$.). We set default context by tokenizing and encoding the context string "A photo of", as in our implementation of the CoOp run. The default class vectors are initialized in a same way using the classnames from the dataset. The training objective for the new weights were the same as the rest of the pipeline, as in the case of our implementation of the CoOp prompt learning strategy.

\noindent \textbf{Target Optimization} is the prompt learning alternative to the proposed POLE pipeline, because essentially it just learns the class vector, keeping the rest of the prompt fixed (`\textit{A photo of} [V$_{CLS}$]."). As in the other methods, the training objectives remains the same as in \autoref{eq:cont}. The class vectors are initialized using the same initialization scheme as DeFo and CoOp experiments.

Please note that as our subsequent downstream task is weakly supervised semantic segmentation, we could not employ the exact model specifications used by \cite{wang2022learning} and \cite{zhou2022conditional} in their entirety. We only used their prompt learning scheme for analyzing what impact these techniques may have on Weakly Supervised Semantic Segmentation. We also found 2.5e-7 experimentally to be a good learning rate for training the prompt weights, which is 0.001 times the learning rate for the rest of the pipeline. Using learning rates larger impacts the performance drastically. The process of handling the text prompts for each of these techniques is described schematically in \autoref{fig:prompt_learning}. The text prompt input it tokenized into a vector of length 77 (Diagram shows only 11 for simplicity). Each token in the string is then tokenized into an unique number that composes the tokenized vector. Typically, there is a number denoting the start of the prompt, called the \textit{prefix}, followed by the \textit{context} tokens, the \textit{class} token, a \textit{punctuator} token (fullstop, in this case) and a blank \textit{suffix} (an array of zeroes). This vector is then embedded as floating point tensors where each token is turned into a vector of fixed length. We selectively convert these vectors as trainable parameters or frozen, depending upon what prompt learning scheme we replicate. Eventually, the CLIP text encoder encodes the embedding and generates a tensor of the same dimensions as that of the masked image encoding. 

\subsection{Additional details on the corpus and synonyms}
\label{sup:ablation-corpus}

In our method, we employ the synonyms obtained from the ChatGPT training corpus. As this corpus is not formally accessible, we use the chatGPT web application to provide synonyms to the ground truth categories. We used the input query prompt: \textit{"Give me 5 semantically similar words for [CLS] and also print the cosine similarity scores based on CLIP model"}; where [CLS] is a classname. From the list of synonyms obtained from GPT, top $m$ synonyms for each class were taken. The list of synonyms collected from the ChatGPT web application are listed in \autoref{list_syn}. 

\begin{table}[h!]
\caption{List of synonyms for each class (associated PascalVOC ground truth) obtained from ChatGPT.}
\label{list_syn}
\resizebox{\columnwidth}{!}{%
\begin{tabular}{cccc}
\hline
\textbf{Class} & \textbf{Synonym 1} & \textbf{Synonym 2} & \textbf{Synonym 3} \\ \hline
aeroplane      & aircraft           & airplane           & plane              \\
bicycle        & bike               & cycle              & pedal bike         \\
bird           & avian              & fowl               & feathered friend   \\
boat           & ship               & vessel             & watercraft         \\
bottle         & flask              & container          & jar                \\
bus            & coach              & transit            & omnibus            \\
car            & automobile         & vehicle            & sedan              \\
cat            & feline             & kitty              & tomcat             \\
chair          & seat               & armchair           & recliner           \\
cow            & bovine             & heifer             & bull               \\
dining table   & kitchen table      & dinner table       & breakfast table    \\
dog            & canine             & puppy              & hound              \\
horse          & equine             & mare               & stallion           \\
motorbike      & motorcycle         & bike               & scooter            \\
player         & person             & individual         & human              \\
potted plant   & houseplant         & flowerpot          & planter            \\
sheep          & lamb               & ewe                & ram                \\
sofa           & couch              & loveseat           & settee             \\
train          & railway            & locomotive         & subway             \\
tv monitor     & television         & display screen     & flat screen        \\ \hline
\end{tabular}%
}
\end{table}

Furthermore, to evaluate the performance of using synonyms based on the popular ChatGPT, we trained our model based on four different corpus: English Wikipedia, Google News, British National Corpus and English Gigaword. For each class, we search for the 10 closest words to the ground truth class name on the webvectors service (\href{http://vectors.nlpl.eu/explore/embeddings/en/associates/}{http://vectors.nlpl.eu/explore/embeddings/en/associates/}). We then selected the top $m$ words from the list for every class in each corpus. The similarity scores are based on the word2vec models employed by the web application, trained on the corpuses mentioned. For each corpus, we use the ground truth category name jointly with the top-$m$ synonyms for every class, in order to construct a pool of words from which CLIP selects the closest word to a given masked image.

\begin{figure*}[t!]
    \centering
    \includegraphics[width=0.95\textwidth]{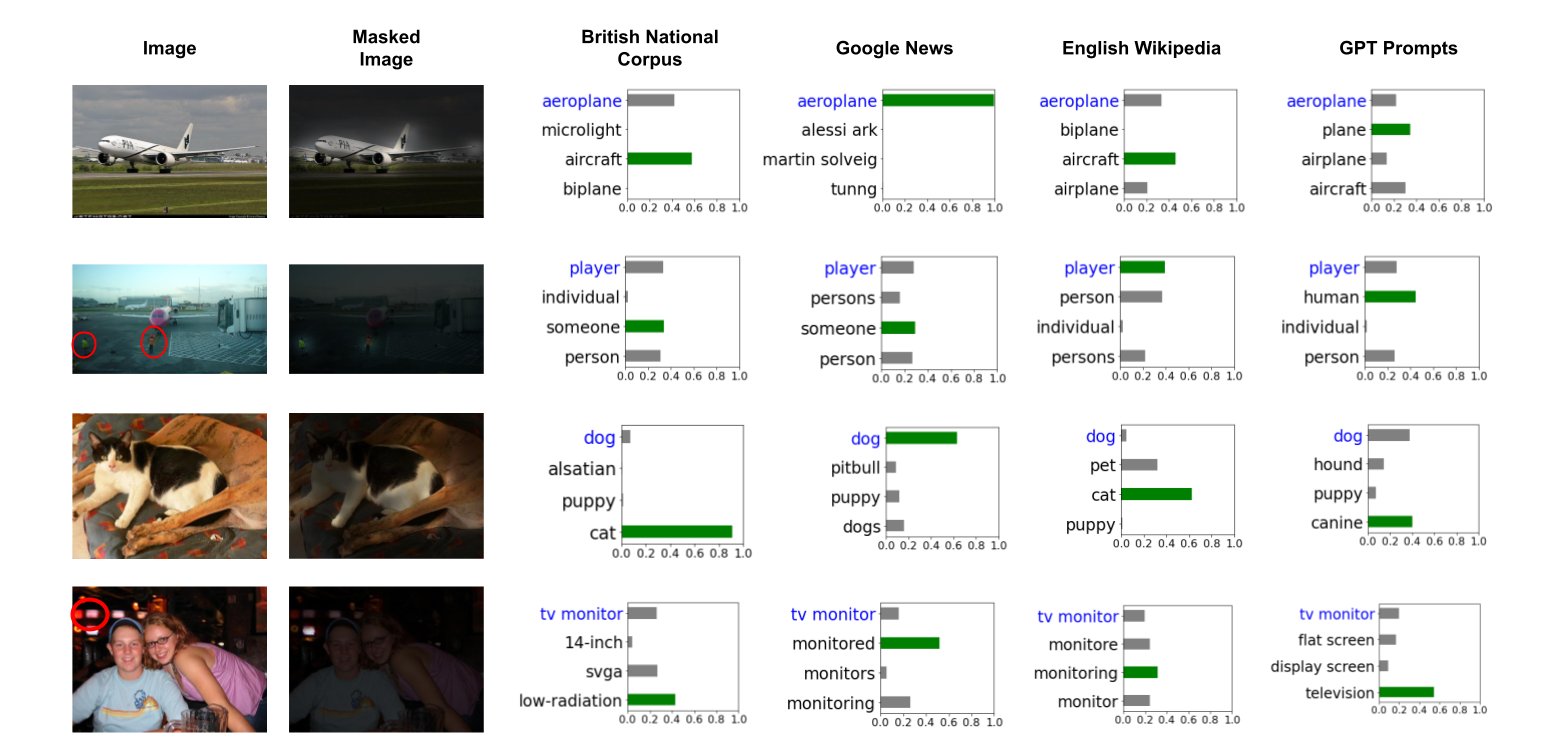}
    \caption{\textbf{Which is the best synonym across corpus?} This figure illustrates several examples of the best synonym selected (\textit{green bar}) for the different corpus, compared to the associated image ground truth (\textit{blue font}). We use red circles to identify the target class.}
    \label{fig:corpus}
\end{figure*}

In Figure \ref{fig:corpus} we further investigate, for several images, which is the best synonym selected across the different corpus. We can observe, for example, that for most corpus the associated ground truth category to a given image is rarely selected. While these correlations may come from the subjectivity when describing an object (e.g., \textit{`aeroplane" vs. `plane"}, or \textit{`tv monitor" vs.`television"}), we believe that in other cases the problem is magnified due to suboptimal category descriptions. Particularly, the class \textit{`person"} is employed systematically in \pascal, which it is replaced in CLIMS \cite{xie2022clims} by \text{`player"}. Nevertheless, we observe that even having both class names in the list of potential synonyms, none of them present the highest correlation for a given image (i.e., three out of four corpus select other category names: \textit{`someone"}, \textit{`someone"} and \textit{`human"}).

\begin{figure*}[t!]
    \centering
    \includegraphics[width=0.95\textwidth]{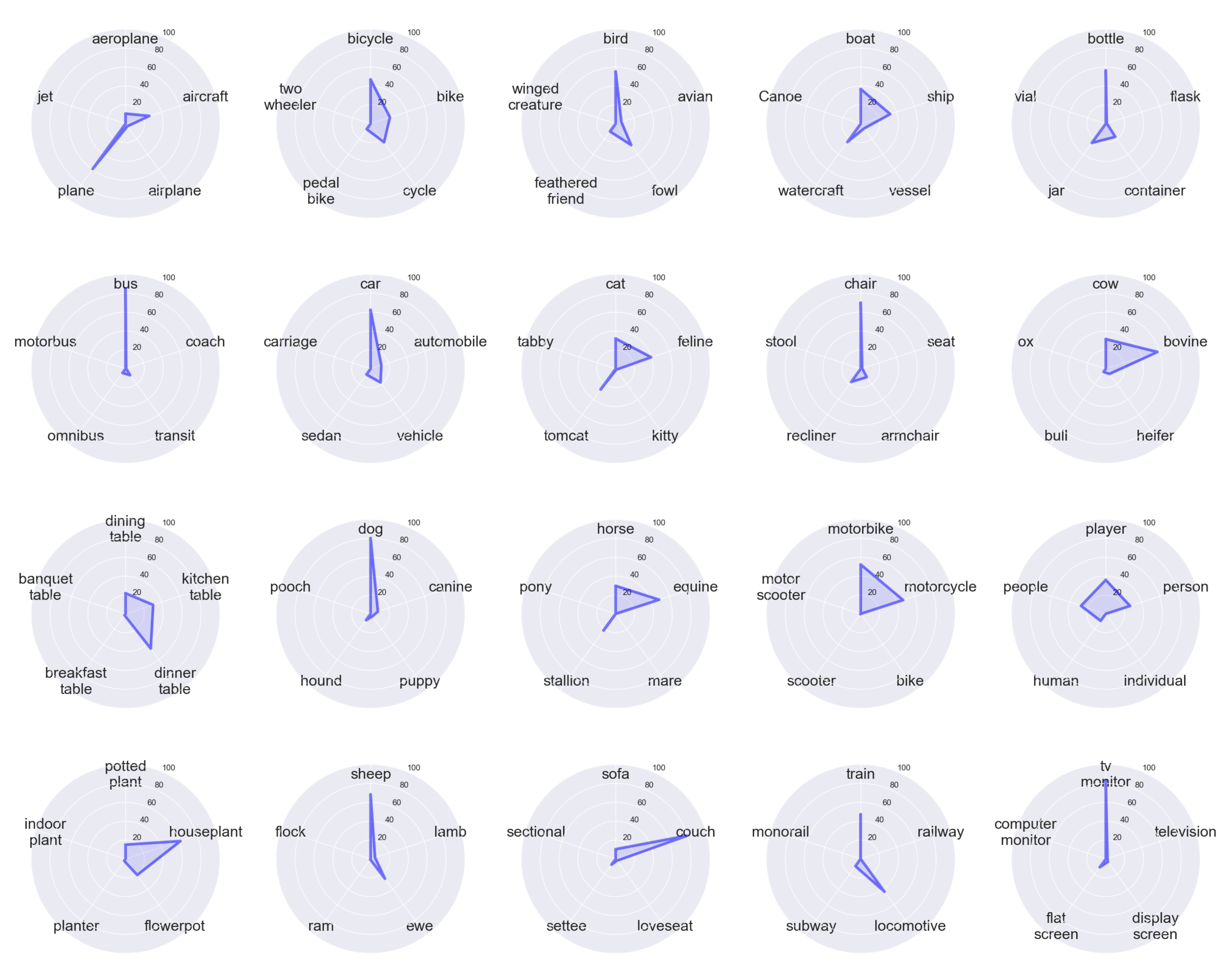}
    \caption{Classwise radial plots for respective fractions of the total number of instances, where a certain word was chosen for an instance of the class. An inward point on the plots indicates that the number of instances where the primary classname itself was chosen in the best prompt, is quite low.}
    \label{fig:clip_selection}
\end{figure*}

This analysis is further supported by the radar plots in Figure \ref{fig:clip_selection}, which depict the frequency at which each ground truth category is selected as the [CLS] token. These plots reveal interesting observations, which suggest that only three classes (i.e., \textit{`bus"}, \textit{`dog"} and \textit{`TV monitor"}) are indeed the most correlated categories in more than 80\% of the images of the whole dataset. On the other hand, a vast majority of categories, the ground truth labels are not selected as optimal synonym in at least 50\% of the images.

\end{document}